\documentclass[twocolumn,10pt,final]{asme2ej}

\usepackage{subfigure}
\usepackage{color}
\usepackage{psfrag}

\usepackage{pstricks}

\usepackage{bm}          % to make bold greek letters in math mode

\usepackage{algorithm} 
\usepackage{algorithmic} 

\usepackage{amsmath,amssymb,epsfig}     % for the use of formule alignment \begin{align} \end{align}

\DeclareMathOperator{\sat}{sat}

\title{Cusp Points in the Parameter Space of Degenerate 3-R\underline{P}R Planar Parallel Manipulators}

\author{Montserrat Manubens
     \affiliation{
	\small Institut de Rob\`otica i Inform\`atica Industrial,\\
	\small CSIC - UPC,\\
	\small Llorens i Artigas, 4-6,\\
	\small 08028 Barcelona, Spain\\
	\small  mmanuben@iri.upc.edu
    }
}

\author{Guillaume Moroz
    \affiliation{
	\small INRIA Nancy-Grand Est,\\
	\small 615, rue du jardin botanique,\\
	\small 54600, Villers-l\`es-Nancy, France\\
	\small guillaume.moroz@inria.fr
	}
}

\author{Damien Chablat
    \affiliation{
	\small Institut de Recherche en Communications \\
        \small et Cybern\'etique de Nantes,\\
	\small UMR CNRS n 6597,\\
	\small 1 rue de la No\"e,\\
	\small 44321 Nantes, France\\
	\small damien.chablat@irccyn.ec-nantes.fr
    }	
}

\author{Philippe Wenger
    \affiliation{
	\small Institut de Recherche en Communications \\
	\small et Cybern\'etique de Nantes,\\
	\small UMR CNRS n 6597,\\
	\small 1 rue de la No\"e,\\
	\small 44321 Nantes, France\\
	\small philippe.wenger@irccyn.ec-nantes.fr
	}	
}

\author{Fabrice Rouillier
    \affiliation{
	\small INRIA Paris-Rocquencourt,\\
	\small Universit\'e Pierre et Marie Curie Paris VI\\
	\small 4, place Jussieu,\\
	\small F-75005 Paris, France\\
	\small fabrice.rouillier@inria.fr
	}	
}

\begin{document}

\maketitle    

%%%%%%%%%%%%%%%%%%%%%%%%%%%%%%%%%%%%%%%%%%%%%%%%%%%%%%%%%%%%%%%%%%%%%%
\begin{abstract}
{\it This paper investigates the conditions in the design parameter space for
the existence and distribution of the cusp locus for planar parallel
manipulators.  Cusp points make possible non-singular assembly-mode changing
motion, which increases the maximum singularity-free workspace. An accurate
algorithm for the determination is proposed amending some imprecisions done by
previous existing algorithms. This is combined with methods of Cylindric
Algebraic Decomposition, Gr\"obner bases and Discriminant Varieties in order to
partition the parameter space into cells with constant number of cusp points.
These algorithms will allow us to classify a family of degenerate
3-R\underline{P}R manipulators. 
}
\end{abstract}

%%%%%%%%%%%%%%%%%%%%%%%%%%%%%%%%%%%%%%%%%%%%%%%%%%%%%%%%%%%%%%%%%%%%%%

%%%%%%%%%%%%%%%%%%%%%%%%%%%%%%%%%%%%%%%%%%%%%%%%%%%%%%%%%%%%%%%%%%%%%%%%%
%%%%%%%%%%%%%%%%%%%%%%%%%%%%%%%%%%%%%%%%%%%%%%%%%%%%%%%%%%%%%%%%%%%%%%%%%
\section{Introduction}
%%%%%%%%%%%%%%%%%%%%%%%%%%%%%%%%%%%%%%%%%%%%%%%%%%%%%%%%%%%%%%%%%%%%%%%%%
%%%%%%%%%%%%%%%%%%%%%%%%%%%%%%%%%%%%%%%%%%%%%%%%%%%%%%%%%%%%%%%%%%%%%%%%%
In the past, singularities were believed to physically separate the different
assembly modes, meaning that for fixed joint values one could not find a path
going from one assembly mode to another without crossing a singular
configuration. So the interest relied on considering the widest connected
non-singular domain, called \emph{aspect}. Innocenti and Parenti-Castelli 
pointed out in  \cite{IPC98} that non-singular changes of assembly mode are
possible, and McAree and Daniel showed in~\cite{McaD99} that such changes are
possible when triple roots of the Forward Kinematic Problem (FKP) exist.
In~\cite{ZWCr07} Zein, Wenger and Chablat showed that for the case of
3-R\underline{P}R manipulators a non-singular change of assembly mode can be
accomplished by encircling a cusp point, and Husty recently proved
in~\cite{Hck09} that the generic 3-R\underline{P}R parallel manipulators without
joint limits always have 2 aspects. 

From the algebraic point of view, the locus of cusp points can be described by
means of symbolic equations. In order to avoid long symbolic-algebraic
manipulations, these equations are usually solved by numerical approximation at
an early stage, which may lead to small deviations that can be propagated along
the process.  However, there exist efficient symbolic-algebraic techniques that
may leave the use of numerical methods to the last step. In particular, we will
apply \emph{Gr\"obner bases}~\cite{CoCoA1}  in order to adopt a more suitable
equivalent system defining the same solution points. 

Lazard and Rouillier introduced the mathematical notion of \emph{Discriminant
Variety} (DV)~\cite{LR07}, which is a variety of codimension 1 in the chosen
parameter space whose complement satisfies the property that over each connected
component the given system has a constant number of solutions.   The complement
of this DV will be partitioned into cells by a \emph{Cylindric Algebraic
Decomposition}~\cite{Cbook75},  also known as CAD.  

This paper is intended to illustrate both the performance of the new algorithm
for the determination of the locus of cusp points and its combination with the
forementioned algebraic techniques in the  analysis of existence conditions and
distribution along a 2-dimensional parameter  space. Although the method can be
applied to more general manipulators (see \cite{ICRA2011}), such performance
will be exemplified on a family of degenerate 3-R\underline{P}R manipulators,
detailed in Section~\ref{sec_3rpr}. The algorithm for the cusp point
determination, which is one of the main contributions of the paper, is given in
Section~\ref{sec_cusps}, where it is compared to other previous algorithms.
Section~\ref{sec_1dimdiscussion} outlines some of the exploited algebraic
objects such as the DV.  In Section~\ref{sec_CAD} the previous procedures are
combined with the CAD to partition a 2-dimensional space with regard to the
associated number of cusp points, which leads us to analyze a complete family of
degenerate 3-R\underline{P}R manipulators that depend on one geometric
parameter. This section also illustrates some  applications of the presented
strategy to robot design.    The paper concludes in
Section~\ref{sec_conclusion}.

%%%%%%%%%%%%%%%%%%%%%%%%%%%%%%%%%%%%%%%%%%%%%%%%%%%%%%%%%%%%%%%%%%%%%%%%%
%%%%%%%%%%%%%%%%%%%%%%%%%%%%%%%%%%%%%%%%%%%%%%%%%%%%%%%%%%%%%%%%%%%%%%%%%
\section{A class of degenerate 3-R\underline{P}R  \label{sec_3rpr}}
%%%%%%%%%%%%%%%%%%%%%%%%%%%%%%%%%%%%%%%%%%%%%%%%%%%%%%%%%%%%%%%%%%%%%%%%%
%%%%%%%%%%%%%%%%%%%%%%%%%%%%%%%%%%%%%%%%%%%%%%%%%%%%%%%%%%%%%%%%%%%%%%%%%
Let us describe the family of manipulators on which the strategies presented
along the paper will be exemplified.  A general 3-R\underline{P}R manipulator is
a 3-degrees-of-freedom planar parallel mechanism that has two platforms
connected by   three R\underline{P}R rods, with the prismatic joints being
actuated and the revolute ones being passive. Without loss of generality we can
assume the absolute reference frame to be such that the base points of the leg
rods are $A_1=(0,0)$, $A_2=(A_{2x},0)$ with $A_{2x} > 0$, and $A_3=(A_{3x},
A_{3y})$. If $B_1$, $B_2$ and $B_3$ are the corresponding points on the moving
platform, then  the geometric parameters associated to this manipulator are the
values $A_{2x}$, $A_{3x}$, $A_{3y}$, the lengths $d_1=\|\overline{B_1B_2}\|$,
$d_3=\|\overline{B_1B_3}\|$, and the angle $\beta=\widehat{B_2B_1B_3}$. The
input-space is then formed by $\bm\rho=(\rho_1,\rho_2,\rho_3)\in \mathbb{R}^3$,
where $\rho_i\geq 0$ are the leg rod lengths, and the output-space is formed by
the poses of the moving platform $\mathbf{x}=(x,y,\alpha)$, where $B_1=(x,y)$
and $\alpha$ is the angle of vector $B_2-B_1$ relative to $A_2-A_1$. We define
$(s_{\alpha}, c_{\alpha})$, $(s_{\beta}, c_{\beta})$ and $(s_{\alpha + \beta},
c_{\alpha + \beta})$ to denote the sines and cosines of $\alpha$, $\beta$, and
$(\alpha + \beta)$, respectively. Then the forward kinematics of a general
3-R\underline{P}R manipulator is defined by the system of equations  
\begin{align}
\begin{array}{rl} 
    x^2 + y^2 - \rho_1^2 &=0\\
   (x + d_1 \, c_{\alpha} - A_{2x})^2 + (y + d_1 \, s_{\alpha})^2 - \rho_2^2 &=0\\
   (x + d_3 \, c_{\alpha + \beta} - A_{3x})^2 + (y + d_3 \, s_{\alpha + \beta} - A_{3y})^2 - \rho_3^2 &=0.
\end{array}
\label{eq_3RPR}
\end{align}

\begin{figure}[t]
\centering
   \psfrag{A1}{$A_1$}
   \psfrag{A2}{$A_2$}
   \psfrag{A3}{$A_3$}
   \psfrag{B1}{$B_1$}
   \psfrag{B2}{$B_2$}
   \psfrag{B3}{$B_3$}
   \psfrag{a}{$\alpha$}
   \psfrag{b}{$\beta$}
   \psfrag{r1}{$\rho_1$}
   \psfrag{r2}{$\rho_2$}
   \psfrag{r3}{$\rho_3$}
   \psfrag{d1}{$d_1$}
   \psfrag{d3}{$d_3$}
   \psfrag{x}{$x$}
   \psfrag{y}{$y$}
   \includegraphics[width=\linewidth ]{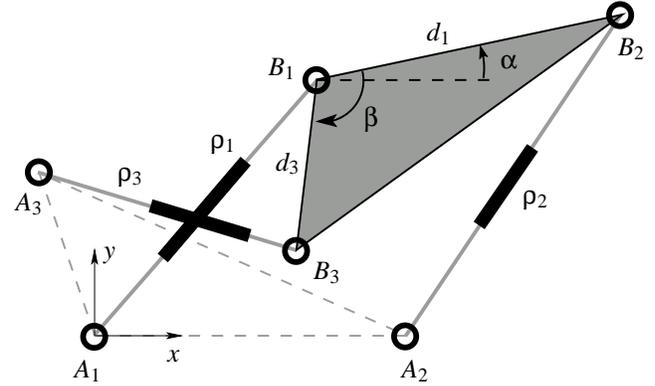}  
   \caption{Example of degenerate 3-R\underline{P}R.}
\label{Fig_Degenerate3RPR}
\end{figure}

For these manipulators, Hunt showed that the FKP admits at most 6 assembly
modes~\cite{Hunt83}, and several authors~\cite{GRSmmt92,PennKass90} proved
independently that the system associated to the FKP can be reduced to  a
polynomial of degree 6.  The 3-R\underline{P}R manipulators for which the degree
of this characteristic polynomial decreases are known as \emph{analytic} or
\emph{degenerate}~\cite{KonGos01,WenChaCK09}, because the Cramer system in
Gosselin's method degenerates.  In this paper we will focus on a class of
degenerate 3-R\underline{P}R manipulators  whose base and moving platforms are
congruent triangles, with the moving triangle being reflected with respect to
the base one, as that of Fig.~\ref{Fig_Degenerate3RPR}.   This class of
manipulators was first studied by Wenger, Chablat and Zein
in~\cite{WenChaZei07}. Their mathematical description requires the addition to
the initial Eqn.~\eqref{eq_3RPR} of the following geometric constraints: 
\begin{align}
\begin{array}{rl} 
  d_1 =& A_{2x}\\
  \cos(\beta) =&  A_{3x}/d_3 \\ 
  \sin(\beta) =& -A_{3y}/d_3.
\end{array}
\label{eq_3RPRdeg} 
\end{align}
Therefore, the system of equations defining this family of degenerate
3-R\underline{P}R manipulators, formed by Eqns.~\eqref{eq_3RPR} and
~\eqref{eq_3RPRdeg}, will be denoted as  
\[
  F(\bm{\rho},\mathbf{x})=0.
\]
Generically, we will refer to this system by $F$. Whenever concrete values for
the geometric parameters are considered, these will be specified. Finally, the
notation $|_{(\bm{\rho},\mathbf{x})}$ will stand for the evaluation on real
values $(\bm{\rho},\mathbf{x})$.

%%%%%%%%%%%%%%%%%%%%%%%%%%%%%%%%%%%%%%%%%%%%%%%%%%%%%%%%%%%%%%%%%%%%%%%%%
%%%%%%%%%%%%%%%%%%%%%%%%%%%%%%%%%%%%%%%%%%%%%%%%%%%%%%%%%%%%%%%%%%%%%%%%%
\section{Cusp locus determination \label{sec_cusps}}
%%%%%%%%%%%%%%%%%%%%%%%%%%%%%%%%%%%%%%%%%%%%%%%%%%%%%%%%%%%%%%%%%%%%%%%%%
%%%%%%%%%%%%%%%%%%%%%%%%%%%%%%%%%%%%%%%%%%%%%%%%%%%%%%%%%%%%%%%%%%%%%%%%%
In this section we describe the cuspidal locus and analyze the usual methods for
their determination. After that we propose a more accurate approach and compare
it to the previous ones in a simple example.  

\begin{figure}[t]
\centering
   \psfrag{p1}{$p_1$}
   \psfrag{p2}{$p_2$}
   \psfrag{p3}{$p_3$}
   \psfrag{k}{$\kappa$}
   \psfrag{Pip}{$\pi_{\rho}(p_i)$}
   \psfrag{Pik}{$\pi_{\rho}(\kappa)$}
\includegraphics[scale=0.5]{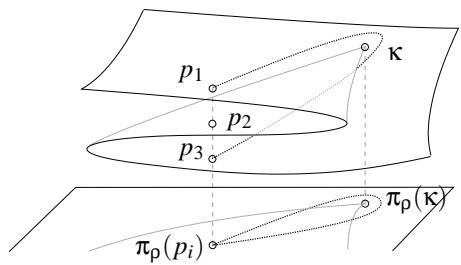}
\caption{Cusp point $\kappa$ as a triple root of the FKP and non-singular path linking upper and lower solutions of the FKP.}
\label{Fig_Cusppoint}
\end{figure}

Let us assume that a specific manipulator, whose geometric parameters have
been set into $F$, has been designated. Then, we denote the associated
\emph{configuration space} by 
\[
\mathcal{C}(F)= \{(\bm{\rho},\mathbf{x}) \in \mathbb{R}^6 \,:\, F|_{(\bm{\rho},\mathbf{x})}=0\}.
\]  
The Jacobian matrix of $F$ with respect to the output variables is denoted as 
$\mathbf{J_{x}}(F) = \left(
\begin{array}{ccc} 
\frac{\partial F}{\partial x} & \frac{\partial F}{\partial y}  & \frac{\partial F}{\partial \alpha}
\end{array} 
\right)$. 
The configurations where its determinant is zero are called \emph{parallel
singular configurations},  or \emph{type 2 singularities}. On these
configurations the manipulator shows a loss of control. The \emph{parallel
singular locus} of our manipulator is a 2-dimensional space that can be
described (see~\cite{GAieee90}) as 
\[ \Sigma(F)= \{(\bm{\rho},\mathbf{x}) \in \mathcal{C}(F) \, : \, \mathbf{J_{x}}(F)|_{(\bm{\rho},\mathbf{x})} \text{ is rank deficient} \}.
\] 
For simplicity, we will refer to this set as the \emph{singular locus}. 
With this setting we now define the \emph{cuspidal locus} as
\[ \kappa(F)= \{(\bm{\rho},\mathbf{x}) \in \mathcal{C}(F) \, : \, \bm{\rho} \text{ root of exact multiplicity 3 of } F\},
\]  
i.e. the triple roots of the FKP.  Observe that $\kappa(F) \subset \Sigma(F)$,
since the Jacobian $\mathbf{J_{x}}(F)$ is rank deficient on the roots of
multiplicity three of $F$. It is known that in the proximity of cusp points a non-singular change
of assembly mode can be made.  Figure~\ref{Fig_Cusppoint} shows a cusp point
$\kappa$ and a non-singular path connecting two different assembly modes ($p_1$
and $p_3$). We shall note, however, that both the singular and the cusp locus
are quite difficult to visualize in the 6-dimensional
$(\bm{\rho},\mathbf{x})$-space. So for mechanisms with at most one inverse
kinematics solution, as is the case for the 3-R\underline{P}R, we will actually
project them onto the input-space instead. 

%%%%%%%%%%%%%%%%%%%%%%%%%%%%%%%%%%%%%%%%%%%%%%%%%%%%%%%%%%%%%%%%%%%%%%%%%
\subsection{Usual methods for the cusp computation }
%%%%%%%%%%%%%%%%%%%%%%%%%%%%%%%%%%%%%%%%%%%%%%%%%%%%%%%%%%%%%%%%%%%%%%%%%
Let us revise the two main algorithms that have been more commonly used in the
determination of the locus of cusp points for a given manipulator. The following
method, introduced by Wenger and Chablat in~\cite{WenChaCK09}, and anallytically
derived in~\cite{UrizarPetuya2011}, has been used for the degenerate 
$3$-R\underline{P}R manipulators. It was inspired on an approach developed by
Hern\'andez et al in~\cite{HernandezAltuzarra2009} for other robots.
\begin{algorithm}[H]
\caption{by Wenger and Chablat~\cite{WenChaCK09}}
\label{Alg1}
\begin{algorithmic}
\STATE{1. Reduce $F$ (by successive resultants) to a single equation}   
\STATE{\ \hspace{2mm} $g(t)=0$, with $t=\tan(\alpha/2)$ and coefficients in $\bm{\rho}$.} 
\STATE{2. Equations of triple roots of $g$}
\STATE{\ \hspace{2mm} $G=$ $\{g=0 , \frac{\partial g}{\partial t}=0, \frac{\partial^2 g }{\partial t^2}=0\}$.}
\STATE{3. Equations of strictly triple roots of $g$}
\STATE{\ \hspace{2mm} $G=$ $G \cup \{ \frac{\partial^3 g }{\partial t^3}\neq 0  \}$.}
\STATE{4. $\widetilde{G}=$ Eliminate $t$ from $G$ and solve the remaining system for real values of $\bm{\rho}$.}
\STATE{5. Solve $\widetilde{G}$.}
\end{algorithmic}
\end{algorithm}

This strategy reduces the problem to the computation of the strictly triple
roots of one single univariate polynomial $g$. However, the constraint added in
step 3 makes the computation quite hard, and thus this step is often removed. 

Another commonly used method, described by McAree and Daniel in~\cite{McaD99},
makes use of the series expansion of $F$. 
\begin{algorithm}[H]
\caption{by McAree and Daniel~\cite{McaD99} }
\label{Alg2}
\begin{algorithmic}
  \STATE{1. Series expansion of $F$} 
  \STATE{ ${\footnotesize 
\Delta F = \frac{\partial F}{\partial \mathbf{x}} \Delta \mathbf{x} + \frac{\partial F}{\partial \bm{\rho}} \Delta \bm{\rho} + \frac{1}{2} 
  \Delta \mathbf{x}^T \left( 
    \frac{\partial^2 F}{\partial \mathbf{x}^2} \right) \Delta \mathbf{x} +  
  \Delta \mathbf{x}^T \left(
    \frac{\partial^2 F}{\partial \mathbf{x} \partial \bm{\rho}} 
 \right) \Delta \bm{\rho} +}$}
  \STATE{\ \hspace{7mm}  ${\footnotesize\frac{1}{2} \Delta \bm{\rho}^T \left( \frac{\partial^2 F}{\partial \bm{\rho}^2} \right) \Delta \bm{\rho} + \dots }$ }
  \STATE{2. Compute configurations where 1st and 2nd order constraints are rank deficient, i.e. solve}
  \STATE{\ \hspace{1mm} $\mathbf{v}^T \left( \mathbf{u} \frac{\partial^2 F}{\partial \mathbf{x}^2}\right) \mathbf{v} = 0$, where $\mathbf{v}$ is a unit vector in right kernel}
  \STATE{\ \hspace{1mm} of $\frac{\partial F}{\partial \mathbf{x}}$, and $\mathbf{u}$ is a unit vector that spans left kernel. }
\end{algorithmic}
\end{algorithm}

This second strategy reduces the problem to the resolution of some quadratic
equations, but it also requires to find the unit vectors $\mathbf{u}$ and
$\mathbf{v}$, which may hinder the computation.

These algorithms are commonly used in the cusp locus determination. However,
both have drawbacks related to the non-cuspidality of some resulting points: 
\begin{itemize}
\item[$\bullet$] 
Since the polynomial $g$ obtained by Algorithm~\ref{Alg1} is the result of
several projections, some of the obtained points may correspond to the
projection of complex (not real) solutions, as we will see later on. 
\item[$\bullet$] Step 3 usually needs to be removed from Algorithm~\ref{Alg1} in
order to avoid slow-processing. 
\item[$\bullet$] Algorithm~\ref{Alg2} does not constrain the multiplicity of the
solutions to be exactly 3, so it may obtain higher multiplicity ones. Regardless
of that, in~\cite{ZWCr07} it is shown that additional spurious solutions may be
produced for generic 3-R\underline{P}R manipulators.   
\end{itemize}
Therefore, both methods can only provide sufficient conditions for the cuspidal
locus but not always necessary ones. 

%%%%%%%%%%%%%%%%%%%%%%%%%%%%%%%%%%%%%%%%%%%%%%%%%%%%%%%%%%%%%%%%%%%%%%%%%
\subsection{Improved method}
%%%%%%%%%%%%%%%%%%%%%%%%%%%%%%%%%%%%%%%%%%%%%%%%%%%%%%%%%%%%%%%%%%%%%%%%%
Despite the fact that the formulation of the cusp locus is quite simple, the associated system
of equations usually contains many equations in many unknowns, whose resolution
can take long computations and even lead to abnormal termination for not too
complex examples. So the methods described previously were introduced as simple,
though not accurate, alternatives to the symbolic resolution. However, we can
now get over some of these difficulties with current powerful symbolic algebra
tools that fix the deficiencies of the algorithms detailed above. 

The approach that we propose  is an evolution of
~\cite{MorozRouillierChablatWenger2010} by Moroz et al., inspired on the results
of~\cite{MorozThesis}. The main difference of the proposed method compared to
that of~\cite{MorozRouillierChablatWenger2010} is the introduction of the
\emph{saturation} operator to remove the quadruple roots. 

\begin{algorithm}[H]
\caption{Proposed method}
\label{Alg3}
\begin{algorithmic}
\STATE{1. Equations of double roots of $F$ w.r.t. $\bm{\rho}$}
\STATE{\ \hspace{2mm} $D_{F}= F \cup \{ \det(\mathbf{J_{x}}(F)) =0 \}$ }
\STATE{2. Equations of triple roots of $F$ w.r.t. $\bm{\rho}$}
\STATE{\ \hspace{2mm} $T_{F}= D_{F} \cup \{\det(\mathbf{m})\,:\, \mathbf{m}$ maximal minors of $\mathbf{J_{x}}(D_{F}) \}$ } 
\STATE{3. Equations of quadruple roots of $F$ w.r.t. $\bm{\rho}$}
\STATE{\ \hspace{2mm} $Q_{F}= T_{F} \cup \{\det(\mathbf{m})\,:\, \mathbf{m}$ maximal minors of $\mathbf{J_{x}}(T_{F}) \}$ } 
\STATE{4. Saturate $T_{F}$ by $Q_{F}$} 
\STATE{\ \hspace{2mm} $C_{F}= \sat(T_{F}, Q_{F})$ }
\STATE{5. Solve $C_{F}$ for real values of $(\bm{\rho},\mathbf{x})$}
\end{algorithmic}
\end{algorithm}

Given the system defining the mechanism $F$, it computes iteratively the
equations $T_{F}$ and $Q_{F}$ of triple and quadruple roots $\bm{\rho}$ of $F$,
respectively.  Then, we use saturation. Given two polynomial systems $S_1$ and
$S_2$, $\sat(S_1,S_2)$ is an algebraic operator that returns a polynomial system
whose solution set is the closure of the solutions of the first system after
removing those of the second one. If $V(S_i)$ denotes the solution set of $S_i$,
it is satisfied that  
\begin{equation}\label{Eq_saturation}
\overline{V(\sat(S_1,S_2))} = \overline{V(S_1) \setminus V(S_2)}.
\end{equation}
In general, the saturation ensures that all roots of $S_2$ are removed. However,
in specific cases, some points can remain due to property of
Eqn.~\eqref{Eq_saturation} for which we can only obtain $\overline{V(S_1)
\setminus V(S_2)}$ instead of $V(S_1) \setminus V(S_2)$, which can differ by a
null-measure set that can easily be removed afterwards. Further details on the
saturation and its geometric interpretation can be found in~\cite{CoCoA1}.

Although we are only interested in real (feasible) solutions, we shall note that
the polynomial system obtained after saturating has real coefficients and thus
its solution set could contain some complex (not real) roots. For this reason we
need to solve the final cusp system $C_{F}$ in the real field. This is done by
using the \emph{RootFinding} Maple package.

With Algorithm~\ref{Alg3} the previous drawbacks are amended:
\begin{itemize}
  \item[$\bullet$] When computing the saturation of $T_{F}$ by $Q_{F}$, the
points of multiplicity 4 or higher are removed, and so we can guarantee that
only the cusp locus is obtained.   
  \item[$\bullet$] By solving $C_{F}$ for $(\bm{\rho},\mathbf{x})$, instead
projecting onto the $\bm{\rho}$-space, we avoid having biased points produced by
the projection of complex (not real) solutions. 
  \item[$\bullet$] Furthermore, solving $C_{F}$ in the real field ensures that
no other spurious complex solutions are considered.
\end{itemize}

%%%%%%%%%%%%%%%%%%%%%%%%%%%%%%%%%%%%%%%%%%%%%%%%%%%%%%%%%%%%%%%%%%%%%%%%%
\subsection{Case study comparison}\label{subsec_CaseStudyDegenerate} 
%%%%%%%%%%%%%%%%%%%%%%%%%%%%%%%%%%%%%%%%%%%%%%%%%%%%%%%%%%%%%%%%%%%%%%%%%
Let us now compare the performance of both Algorithm~\ref{Alg1} without step 3
and the proposed Algorithm~\ref{Alg3} on a simple case of degenerate
3-R\underline{P}R in order to contrast their results. However, let us clarify
that both the formulation and the proposed algorithm apply to other more general
manipulators (see \cite{ICRA2011}). We set the geometric parameter values
$A_{2x}=1$, $A_{3x}=0$, $A_{3y}=1$, $\beta=-\pi/2$, $d_1=1$, and $d_3=1$.

The characteristic polynomial for Algorithm~\ref{Alg1} is  
\[
g(t)=(\rho_3^2-\rho_1^2)\,t^3 + (\rho_2^2-\rho_1^2-4)\,t^2 + (\rho_3^2-\rho_1^2-4)\,t + \rho_2^2-\rho_1^2
\]
After eliminating $t$ from $G$ we get $\widetilde{G}=\{P_1, P_2, P_3\}$ as follows  
\begin{equation}\label{Eq_G}
\begin{split}
\rho_2^4+\rho_3^4-2\,\rho_2^2\,\rho_3^2+6\,\rho_1^2-3\,\rho_2^2-3\,\rho_3^2 -12 &=0 \\    
2\,\rho_1^4+2\,\rho_3^4-4\,\rho_1^2\,\rho_3^2+4\,\rho_1^2+3\,\rho_2^2-7\,\rho_3^2 -16 &=0 \\
\rho_3^4+\rho_1^2\,\rho_2^2-\rho_1^2\,\rho_3^2-\rho_2^2\,\rho_3^2+3\,\rho_1^2+\rho_2^2-4\,\rho_3^2 -6 &=0.
\end{split}\end{equation} 
Observe that these equations are not independent. Indeed, $P_2$ is a
combination of the other two: 
\[
 P_2= \frac{(\rho_{1}^2-\rho_{3}^2+1)}{3}\,P_1 + \frac{(\rho_{3}^2-\rho_{2}^2+6)}{3}\,P_3.
\]
Additionally, there are solutions of $\widetilde{G}$ that do not correspond to
the real cusp locus. For instance, if we set $\rho_1=1/3$,  the system
$\widetilde{G}|_{\rho_1=1/3}$ has two solutions with both $\rho_2$ and $\rho_3$
positive. But the FKP evaluated on these two solutions  only has complex
solutions $(x,y, \alpha)$. So, for $\rho_1=1/3$ the c-space $\mathcal{C}(F)$ has
no cusp points, though Algorithm 1 obtained two mistaken candidates.

We now test Algorithm~\ref{Alg3} on the same example.  The equations of the cusp
locus $C_{F}$ are:

\begin{equation}\label{Eq_cusps}
\begin{split}
6\,c_{\alpha} + \rho_2^2-\rho_3^2&=0 \\
2\,s_{\alpha}^2 + s_{\alpha} - 1&=0 \\  
2\,c_{\alpha}^2 - s_{\alpha} - 1 &=0 \\  
2\,c_{\alpha}\,s_{\alpha} - c_{\alpha}&=0 \\   
3\,c_{\alpha} + 3\,s_{\alpha} + \rho_1^2 -\rho_3^2 +  1 &=0 \\  
3\,c_{\alpha} + 3\,s_{\alpha} + x^2+y^2 - \rho_3^2 + 1 &=0 \\  
2\,x\,s_{\alpha} + 2\,y\,s_{\alpha} -4\,s_{\alpha} - x-y +2 &=0 \\  
2\,s_{\alpha}\,\rho_3^2 -3\,c_{\alpha}-6\,s_{\alpha} - 4\,x\,y - x-y &=0 \\  
2\,x\,c_{\alpha} + 2\,y\,s_{\alpha} +c_{\alpha}-3\,s_{\alpha} - 2\,x+ 1 &=0 \\  
2\,y\,c_{\alpha} + 2\,y\,s_{\alpha} +c_{\alpha}-s_{\alpha} - x+y+ 1 &=0 \\  
4\,y\,s_{\alpha}-s_{\alpha} +2\,(c_{\alpha}-1)\,\rho_3^2 + 4\,y^2 -3\,y +x + 1 &=0 \\ 
2\,s_{\alpha}\,(2\,y^2 - 4\,y -1) -3\,c_{\alpha} - 4\,x\,y - 2\,y^2 +3\,y & \\ 
-x +\rho_3^2 -2 &=0 \\ 
6\,s_{\alpha}\,(2\,y +1) + 12\,c_{\alpha} -8\,y^2\,x +8\,y^3 +12\,x\,y & \\ 
+x -3\,y +2\,x\,\rho_3^2 -6\,y\,\rho_3^2 -6\,\rho_3^2 +8  &=0 \\
18\,s_{\alpha}\,(18\,y -1) +36\,c_{\alpha} +32\,y^4 +20\,y^2 +44\,x\,y & \\ 
+13\,x -47\,y + 4\,\rho_3^4 - 8\,\rho_3^2\,y\,x - 32\,y^2\,\rho_3^2  & \\
-6\,x\,\rho_3^2 -6\,y\,\rho_3^2-32\,\rho_3^2+ 40 &=0.
\end{split}
\end{equation}
Here if we try to solve $C_F|_{\rho_1=1/3}$,  we get no real solutions.  
\begin{figure}[t] 
\centering
   \psfrag{r2}{$\rho_2$}
   \psfrag{r3}{$\rho_3$}
 \includegraphics[scale=.25]{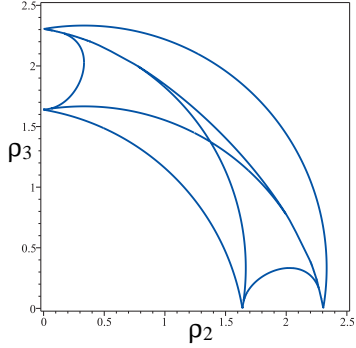}
 \caption{Singular curve for $\rho_1=\frac{1}{3}$ on $(\rho_2,\rho_3)$}\label{Fig_Singcurve}
 \label{Fig_SingCurveR2R3}
\end{figure}
Figure~\ref{Fig_SingCurveR2R3} shows a section in the $(\rho_2, \rho_3)$-plane
of the singular locus of $F$ for $\rho_1=\frac{1}{3}$. As can be observed, 
there is no cusp point in this plot. This phenomenon is not casual. Actually,
the value $1/3$ for $\rho_1$ has not been randomly picked as we will see in next
section. 
In fact, Algorithm~\ref{Alg3} describes the cusp locus more
accurately than Algorithm~\ref{Alg1}, in general. 

%%%%%%%%%%%%%%%%%%%%%%%%%%%%%%%%%%%%%%%%%%%%%%%%%%%%%%%%%%%%%%%%%%%%%%%%%
%%%%%%%%%%%%%%%%%%%%%%%%%%%%%%%%%%%%%%%%%%%%%%%%%%%%%%%%%%%%%%%%%%%%%%%%%
\section{Discussion on the joint space \label{sec_1dimdiscussion}}
%%%%%%%%%%%%%%%%%%%%%%%%%%%%%%%%%%%%%%%%%%%%%%%%%%%%%%%%%%%%%%%%%%%%%%%%%
%%%%%%%%%%%%%%%%%%%%%%%%%%%%%%%%%%%%%%%%%%%%%%%%%%%%%%%%%%%%%%%%%%%%%%%%% 
We now extend our improved method to partition a parameter space with regard to
the associated cusp locus. So, we want to discuss the solutions of a parametric
system.  Among the numerous possible ways of solving parametric systems, we
focus on the use of Discriminant Varieties (DV)~\cite{LR07}   for two main
reasons: it provides a formal decomposition of the parameter space through an
exactly known algebraic variety (no approximation), and it has been successfully
used in similar problems~\cite{CR02}.  

Let us consider a general parametric polynomial system \[\mathcal{F}=\{p_1
(\mathbf{v}) = 0, \ldots, p_m (\mathbf{v}) = 0, q_1 (\mathbf{v}) > 0, \ldots,
q_l (\mathbf{v}) > 0\},\] where $p_1, \ldots, p_m, q_1, \ldots, q_l$ are
polynomials with rational coefficients depending on $\mathbf{v}=(U_1, \ldots,
U_d, X_1, \ldots, X_n)$ with $X_i$ being unknowns and $U_i$ parameters. For
instance, the system describing the cuspidal configurations our manipulator
$C_F$ is parametric if some of the geometric parameters are initially
left free in $F$. The DV associated to system $\mathcal{F}$  is described by a
polynomial equation. This DV  partitions the parameter space into several
regions such that over each open region delimited by the DV  the number of real
solutions of $\mathcal{F}$ is constant.  Prior to defining the DV associated to
$\mathcal{F}$, we need to specify a solver of 0-dimensional systems that will be
used as a black box.

%%%%%%%%%%%%%%%%%%%%%%%%%%%%%%%%%%%%%%%%%%%%%%%%%%%%%%%%%%%%%%%%%%%%%%%%%
\subsection{Basic black-boxes}
%%%%%%%%%%%%%%%%%%%%%%%%%%%%%%%%%%%%%%%%%%%%%%%%%%%%%%%%%%%%%%%%%%%%%%%%%
Let us describe the global solver for 0-dimensional systems that will be used as
a black box in the general algorithm.  We mainly use exact computations, namely
formal elimination of variables (resultants, Gr\"obner bases) and resolution of
0-dimensional systems, including univariate polynomials.

We first compute a Gr\"obner basis of the ideal $\langle p_1, \ldots, p_m
\rangle$ for any ordering, which will help us   detect if the system has or has
not finitely many complex solutions.  If yes, then compute a so called Rational
Univariate Representation (RUR) of $\langle p_1, \ldots, p_m \rangle$ (see
{\cite{Raaecc99}}), which is an equivalent system of the form 
\begin{center}
$\{f (T) = 0, X_1 = \frac{g_1 (T)}{g (T)}, \ldots, X_n =
\frac{g_n (T)}{g (T)} \}$, 
\end{center}
where $T$ is a new variable independent of $X_1, \ldots, X_n$, equipped with a
so called \emph{separating element}  (injective on the solutions of the system)
$u \in \mathbb{Q}[X_1, \ldots, X_n]$ and such that :
$$\begin{array}{ccccc}
  V ( p_1, \ldots, p_m ) & \xrightarrow{u} & V (f) &
  \xrightarrow{u^{-1}} & V( p_1,\ldots,p_m)\\
  (x_1, \ldots, x_n) & \mapsto & \beta=u (x_1, \ldots, x_n) &
  \mapsto & \left( \frac{g_1 (\beta)}{g (\beta)}, \ldots, \frac{g_n (\beta)}{g (\beta)}  \right)
\end{array}$$
defines a bijection between the (real) roots of the system and the (real) roots
of the univariate polynomial $f$. 

We then solve $f=0$, computing so called isolating intervals for its real roots,
i.e. non-overlapping intervals with rational bounds that contain a unique real
root of $f$ (see {\cite{RZ03}}). Finally, interval arithmetic is used in order
to get isolating boxes of the real roots of the system (non-overlapping products
of intervals with rational bounds containing a unique real root of the system),
by studying the RUR over the isolating intervals of $f$. 

In practice, we use the function \emph{RootFinding[Isolate]} from Maple
software, which performs exactly the computations described above.

%%%%%%%%%%%%%%%%%%%%%%%%%%%%%%%%%%%%%%%%%%%%%%%%%%%%%%%%%%%%%%%%%%%%%%%%%
\subsection{Discriminant varieties} \label{subsec_DV}
%%%%%%%%%%%%%%%%%%%%%%%%%%%%%%%%%%%%%%%%%%%%%%%%%%%%%%%%%%%%%%%%%%%%%%%%%
Consider now the constructible set
\begin{center}
  ${\textstyle \mathcal{S} =\{\mathbf{v} \in \mathbb{C}^n : \hspace{0.1em} p_1 (\mathbf{v}) = 0, \ldots, p_m (\mathbf{v}) = 0,}$ \\
  ${\textstyle q_1 (\mathbf{v}) \neq 0, \ldots, q_l (\mathbf{v}) \neq 0\}}$,
\end{center}
and let us assume that for almost all the parameter values this $\mathcal{S}$ is
a finite set of points.  Then, a discriminant variety  of $\mathcal{S}$ with
respect to $(U_1,\ldots,U_d)$ is a variety $\mathcal{V} \subset \mathbb{C}^d$
such that over each connected open set $\mathcal{U}$ not intersecting
$\mathcal{V}$ ($\mathcal{U} \cap \mathcal{V} = \emptyset$), $\mathcal{S}$
defines an analytic covering. In particular, the number of points of
$\mathcal{S}$ over any point of $\mathcal{U}$ is constant.  

Discriminant varieties can be computed using basic and well-known tools from
computer algebra such as Gr\"obner bases {\cite{LR07}}. A full package  is
available in Maple software through the \emph{RootFinding[Parametric]}  package,
which  provides us with a polynomial $DV(\mathcal{S};U_1,\ldots,U_d)$ whose
associated discriminant variety is $\mathcal{V}$.

%%%%%%%%%%%%%%%%%%%%%%%%%%%%%%%%%%%%%%%%%%%%%%%%%%%%%%%%%%%%%%%%%%%%%%%%%
\subsection{Case study comparison} \label{subsec_CaseStudyDiscussion}
%%%%%%%%%%%%%%%%%%%%%%%%%%%%%%%%%%%%%%%%%%%%%%%%%%%%%%%%%%%%%%%%%%%%%%%%%
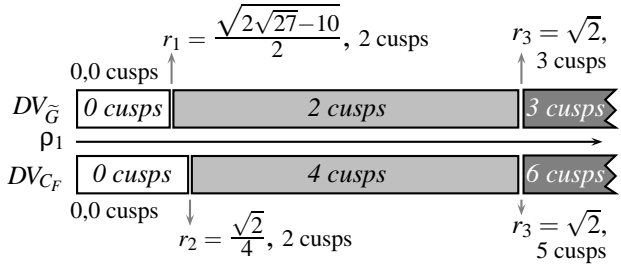
\begin{figure}[t]
\centering
\psset{unit=0.7cm}
\begin{pspicture}(-1.5,-1)(10.5,3.5)
  \rput{0}(-0.75,1.6){$DV_{\widetilde{G}}$}
  \rput{0}(-0.75,0.3){$DV_{C_{F}}$}
  \psframe[fillstyle=solid,fillcolor=white] (0,1.25) (1.8,2)
  \psframe[fillstyle=solid,fillcolor=lightgray] (1.83,1.25) (8.42,2)
  \pspolygon[fillstyle=solid,fillcolor=gray] (8.5,1.27) (8.5,1.99) (10.25,1.99) (10.05,1.8) (10.25,1.6) (10.05,1.4) (10.25,1.27)
  \rput{0}(0.9,1.6){\em 0 cusps}
  \rput{0}(5.15,1.6){\em 2 cusps}
  \rput{0}(9.3,1.6){\textcolor{white}{\em 3 cusps}}
  \psline[linecolor=black]{->}(0,1)(10,1)
  \psframe[fillstyle=solid,fillcolor=white] (0,0) (2.15,0.75)
  \psframe[fillstyle=solid,fillcolor=lightgray] (2.18,0) (8.42,0.75)
  \pspolygon[fillstyle=solid,fillcolor=gray] (8.5,0.02) (8.5,0.74) (10.25,0.74) (10.05,0.55) (10.25,0.35) (10.05,0.15) (10.25,0.02)
    \rput[r]{0}(-0.2,1){$\rho_1$}
    \rput[l]{0}(-0.1,2.3){\small $0$,0 cusps} 
    \rput[l]{0}(-0.1,-0.35){\small $0$,0 cusps} 
    \rput[l]{0}(1.65,3.12){\small $r_1=$ \large $\frac{\sqrt{2\sqrt{27} -10}}{2}$, \small 2 cusps}   
    \psline[linecolor=gray]{->}(1.815,2.10)(1.815,2.7)
    \rput[l]{0}(1.95,-0.75){\small $r_2=$ \large $\frac{\sqrt{2}}{4}$, \small 2 cusps}
    \psline[linecolor=gray]{->}(2.165,-.10)(2.165,-.6)
    \rput[l]{0}(8.28,3.12){$r_3=\sqrt{2}$,} \rput[l]{0}(8.75,2.5){\small 3 cusps}
    \psline[linecolor=gray]{->}(8.46,2.10)(8.46,2.7)
    \rput[l]{0}(8.28,-0.55){$r_3=\sqrt{2}$,} \rput[l]{0}(8.75,-1.1){\small 5 cusps}
    \psline[linecolor=gray]{->}(8.46,-.10)(8.46,-.4)
  \rput{0}(1.07,0.35){\em 0 cusps}
  \rput{0}(5.15,0.35){\em 4 cusps}
  \rput{0}(9.3,0.35){\textcolor{white}{\em 6 cusps}}
\end{pspicture}
   \caption{Comparison of both discussions on $\rho_1$ }
   \label{Fig_discussion_r1}
\end{figure}
Let us consider again the degenerate 3-R\underline{P}R with the same geometric
parameter values as those specified in section~\ref{subsec_CaseStudyDegenerate},
i.e. $A_{2x}=1$, $A_{3x}=0$, $A_{3y}=1$, $\beta=-\pi/2$, $d_1=1$, and $d_3=1$, 
and consider the  systems $\widetilde{G}$ (Eqn.~\ref{Eq_G}), and $C_{F}$
(Eqn.~\ref{Eq_cusps}), obtained by  Algorithm~\ref{Alg1} without step 3 and by
Algorithm~\ref{Alg3}, respectively.  We will regard as a parameter  one of the
leg lengths $\rho_1$ of the manipulator. The discriminant variety will provide
us with a polynomial in $\rho_1$ whose roots will delimit some open intervals
such that for whatever value of $\rho_1$ within one interval, the number
of cusp points of $F|_{\rho_1}$ will be the same.  We compute the DV for each
system with respect to $\rho_1$, and analyze the results. 

In the first case, we get the polynomial  
$DV(\widetilde{G};\rho_1) = \rho_1 \, (\rho_1^2-2) \, (2\rho_1^4+10\rho_1^2-1)$, 
whose roots describing the discriminant variety are  
\begin{center} 
$r_0=0$, \hspace{3mm} $r_1=\frac{\sqrt{2\sqrt{27} -10}}{2}$  
\hspace{2mm} and \hspace{2mm} $r_3=\sqrt{2}$.
\end{center}
Since the number of cusp points is kept constant between two consecutive roots, 
we can compute the associated number of cusps by picking one single value of 
$\rho_1$ inside each open interval and solve $\widetilde{G}|_{\rho_1}$. In 
this case we obtain    
\begin{itemize}
 \item[$\bullet$]$0$ cusp configurations for $\rho_1 \in \, ]0,\,r_1[$,  
 \item[$\bullet$] $2$ cusp configurations for $\rho_1 \in \, ]r_1,r_3[$, and  
 \item[$\bullet$] $3$ for $\rho_1\in \, ]r_3, \infty[$.  
\end{itemize} 
Substituting $\rho_1=r_i$ into $\widetilde{G}$ we obtain the number of cusps on
the borders of the intervals.  
 \begin{itemize}
   \item[$\bullet$] $0$ cusps on $\rho_1=0$, 
   \item[$\bullet$] $2$ cusps on $\rho_1=r_1$, and 
   \item[$\bullet$] $3$ on $\rho_1=r_3$.
 \end{itemize} 

In the second case, a similar analysis for $C_F$ gives 
\[DV(C_{F};\rho_1) = \rho_1 \, (\rho_1^2-2) \, (8\rho_1^2-1) \, (2\rho_1^4+10\rho_1^2-1),\]
which has one more root than $DV(\widetilde{G};\rho_1)$ 
\begin{center}
$r_0=0$, \hspace{3mm} $r_1=\frac{\sqrt{2\sqrt{27} -10}}{2}$, \hspace{3mm} $r_2=\frac{\sqrt{2}}{4}$ \hspace{2mm} and \hspace{2mm} $r_3=\sqrt{2}$.
\end{center}
The intervals and the numbers of cusps for $C_{F}$ differ a bit from those
obtained for $\widetilde{G}$: \vspace{3mm} \\
\begin{minipage}[b]{0.5\linewidth}
\begin{itemize}
 \item[$\bullet$] $0$ cusps for $\rho_1 \in \, ]0,\,r_1[$,
 \item[$\bullet$] $0$ cusps for $\rho_1 \in \, ]r_1,r_2[$,
 \item[$\bullet$] $4$ cusps for $\rho_1 \in \, ]r_2,r_3[$, 
 \item[$\bullet$] $6$ cusps for $\rho_1\in \, ]r_3, \infty[$. 
\end{itemize}
\end{minipage}
\hspace{0.02\linewidth}
\begin{minipage}[b]{0.4\linewidth}
 \begin{itemize}
   \item[$\bullet$] $0$ cusps on $\rho_1=0$, 
   \item[$\bullet$] $0$ cusps on $\rho_1=r_1$, 
   \item[$\bullet$] $2$ cusps on $\rho_1=r_2$, 
   \item[$\bullet$] $5$ cusps on $\rho_1=r_3$.
 \end{itemize} 
\end{minipage}
\vspace{3mm}

The results obtained in both cases are compared in Fig.~\ref{Fig_discussion_r1}.
We can observe that the first two intervals do not exactly coincide, and that
for the second system the obtained numbers of cusps  appear doubled for all
intervals (compared to those obtained for the first system). Both phenomena can
be explained as a consequence of the projection map used to compute the system
$\widetilde{G}$. Let us remind the reader that $\widetilde{G}$ is obtained
after several reductions of the initial system, each of which applying also  a
projection on the $\bm{\rho}$-space. For this, there can be complex
configurations of the manipulator that project onto real roots $\bm{\rho}$ of
$\widetilde{G}$.  This is the case for the values of $\rho_1 \in ]r_1,r_2[$.

The same  can be done for any other parameter and the same phenomena can be
observed. 

%%%%%%%%%%%%%%%%%%%%%%%%%%%%%%%%%%%%%%%%%%%%%%%%%%%%%%%%%%%%%%%%%%%%%%%%%
%%%%%%%%%%%%%%%%%%%%%%%%%%%%%%%%%%%%%%%%%%%%%%%%%%%%%%%%%%%%%%%%%%%%%%%%%
\section{Higher-dimensional discussion by means of a CAD}\label{sec_CAD}  
%%%%%%%%%%%%%%%%%%%%%%%%%%%%%%%%%%%%%%%%%%%%%%%%%%%%%%%%%%%%%%%%%%%%%%%%%
%%%%%%%%%%%%%%%%%%%%%%%%%%%%%%%%%%%%%%%%%%%%%%%%%%%%%%%%%%%%%%%%%%%%%%%% 
By construction we know that over any connected open region not intersecting the
DV the system has a constant number of real roots, for whatever chosen
parameters. But if we want to discuss larger parameter spaces,  then the open
regions will no longer be as simple as 1-dimensional intervals.  So the goal of
this section is to provide an accurate description of the regions with constant
number of solutions. For this we will use the Cylindric Algebraic Decomposition
(CAD)~\cite{Cbook75,DSSissac04}.

%%%%%%%%%%%%%%%%%%%%%%%%%%%%%%%%%%%%%%%%%%%%%%%%%%%%%%%%%%%%%%%%%%%%%%%%%
\subsection{The complementary of a discriminant variety}
\label{section:CAD}
%%%%%%%%%%%%%%%%%%%%%%%%%%%%%%%%%%%%%%%%%%%%%%%%%%%%%%%%%%%%%%%%%%%%%%%%%
Let $\mathcal{P}_d \subset \mathbb{Q}[U_1, \ldots, U_d]$ be the set of
polynomials describing the DV. Then for each $i = d-1, \ldots, 0$, \ we
introduce a new set of polynomials $\mathcal{P}_{i} \subset \mathbb{Q}[U_1,
\ldots, U_{d - i}]$ defined by a backward recursion:
\begin{itemize}
  \item[$\bullet$] $\mathcal{P}_d$ = the polynomials defining the DV,
  \item[$\bullet$] $\mathcal{P}_i$ = $\{$ $DV(p; U_i)$, LeadingCoefficient$(p, U_i)$, 
  \\\indent $\quad\quad$ Resultant$(p, q, U_i)$, $p, q \in \mathcal{P}_{_{i + 1}} \}$
\end{itemize}

Each $\mathcal{P}_i$ has an associated algebraic variety of dimension at most $i
- 1$,  $\mathcal{V}_{i} = V ( \prod_{p \in \mathcal{P}_{i}} p).$  The
$\mathcal{V}_i$ are used to recursively define a finite union of simply
connected open subsets $\cup^{n_{i}}_{k = 1} \mathcal{U}_{i,
k}\subset\mathbb{R}^{i}$ of dimension $i$ such that $V_{i} \cap \mathcal{U}_{i,
k} = \emptyset$.  

Before defining the sets $\mathcal{U}_{i,k}$, we introduce some notation: for a
univariate polynomial $p$ with $n$ real roots,\vspace{2mm}

{\hfill $\operatorname{root}(p,l)=\left\{\begin{array}{l}
    -\infty\mbox{ if }l\leq 0,\\
    \mbox{the $l^{th}$ real root of $p$ if }1\leq l \leq n,\\
    +\infty\mbox{ if }l>n \ . \end{array}\right. $ \hfill} 
\vspace{2mm}\\
Moreover, if $p$ is a $n$-variate polynomial, and $\mathbf{v}$ is a
$(n-1)$-tuple, then $p^\mathbf{v}$ denotes the univariate polynomial where the
first $n-1$ variables have been replaced by $\mathbf{v}$.

The recursive process defining the $\mathcal{U}_{i,k}$ is the following:
\begin{itemize}
 \item[$\bullet$] For $i=1$, let $p_1 = \prod_{p \in \mathcal{P}_1} p$. \\
Taking all $\mathcal{U}_{1, k} =] \operatorname{root}(p_1,k) ;
\operatorname{root}(p_1,k+1) [$ for $k=0,\dots,n$, where $n$ is the number of
real roots of $p_1$, one gets a partition of $\mathbb{R}$ that fits the above
definition. Moreover, one can arbitrarily chose one rational point $u_{1, k}$ in
each open interval $\mathcal{U}_{1, k}$.
\item[$\bullet$] Then, for $i=2,\dots, d$, let $p_i = \prod_{p \in \mathcal{P}_i} p$. \\
The regions $\mathcal{U}_{i,k}$ and the points $u_{i,k}$ are of the form:
$$\begin{array}{l@{}l}
\mathcal{U}_{i,k}=&\left\{(v_1,...,v_{i-1},v_i) \mid \right.\mathbf{v}:=(v_1,...,v_{i-1})\in\mathcal{U}_{i-1,j},\\
&\left. v_i\in ]\operatorname{root}(p_i^{\mathbf{v}},l), \operatorname{root}(p_i^{\mathbf{v}},l+1)[ \right\}\\
u_{i,k}=&(\beta_1,...,\beta_{i-1},\beta_i),\mbox{ with } \\
       & \left\{\begin{array}{l} 
         (\beta_1,...,\beta_{i-1})=u_{i-1,j}\\
          \beta_i\in]\operatorname{root}(p_i^{u_{i-1,j}},l),\operatorname{root}(p_i^{u_{i-1,j}},l+1)[ \ , \end{array} \right.
\end{array}
$$
where $j,l$ are fixed integers.
\end{itemize}

With this recursive procedure we get a full description of the complementary of
the DV  for the system to be solved: the cells $\mathcal{U}_{d, k}$ and a test
point $u_{d,k}\in \mathcal{U}_{d, k}$ (with rational coordinates).   The number
of solutions associated to each open cell $\mathcal{U}_{d, k}$ is obtained by
solving the given system restricted to $u_{d, k}$ using a 0-dimensional solver.
Both the cell decomposition and the test points can be obtained by the Maple
function \emph{RootFinding[Parametric][CellDecomposition]}.

%%%%%%%%%%%%%%%%%%%%%%%%%%%%%%%%%%%%%%%%%%%%%%%%%%%%%%%%%%%%%%%%%%%%%%%%%
\subsection{Open CAD for a class of degenerate 3-R\underline{P}R
\label{SubSec_cells}}
%%%%%%%%%%%%%%%%%%%%%%%%%%%%%%%%%%%%%%%%%%%%%%%%%%%%%%%%%%%%%%%%%%%%%%%%%
\begin{figure}[t]
  \centering
  \psfrag{r1}{$\rho_1$}
  \psfrag{d1}{$d_1$}
  \includegraphics[scale=0.27]{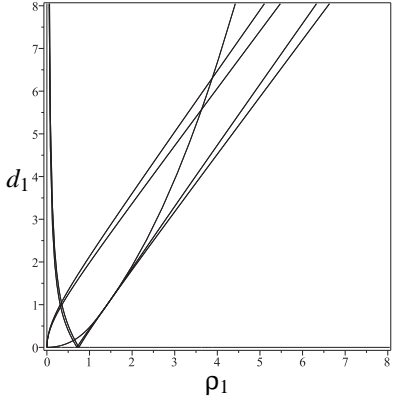}
  \caption{Plot of the DV of $C_F$ with respect to $(\rho_1, d_1)$}
  \label{Fig_DV}
\end{figure}
Let us see the performance of this CAD on a 2-dimensional discussion. We
consider now a family of degenerate 3-R\underline{P}R manipulators with
$A_{2x}=1$, $A_{3x}=0$, $A_{3y}=1$, $\beta=-\pi/2$, and $d_3=1$, and regard as
parameters both $\rho_1$ and $d_1$, constrained by $d_1\geq 0$. Now, the system
$C_{F}$ describing the cusp locus associated to this family of manipulators has
18 polynomials,  its DV is plot in Fig.~\ref{Fig_DV}, and the polynomial
$DV(C_{F};\rho_1, d_1)$ factors as follows: 
\begin{equation*}
\begin{split}
& d_1 \, \rho_1 \, (d_1^2+1) \,(-4\,\rho_1^6-12\,\rho_1^4+27\,\rho_1^2\,d_1^2+15\,\rho_1^2-4) \\
& (4\,\rho_1^6+12\,\rho_1^4\,d_1^2-15\,\rho_1^2\,d_1^4+4\,d_1^6-27\,\rho_1^2\,d_1^2) \\
& (256\,\rho_1^6\,d_1^2+ 81\,\rho_1^2\,d_1^6-288\,\rho_1^4\,d_1^4+256\,\rho_1^6-576\,\rho_1^4\,d_1^2 \\
& +51\,\rho_1^2\,d_1^4 -16\,d_1^4 -288\,\rho_1^4+51\,\rho_1^2\,d_1^2+81\,\rho_1^2).
\end{split}
\end{equation*}

\begin{figure}[t]
\centering
  \psfrag{r1}{$\rho_1$}
  \psfrag{d1}{$d_1$}
  \psfrag{c0}{$0$ cusps}
  \psfrag{c2}{$2$ cusps}
  \psfrag{c4}{$4$ cusps}
  \psfrag{c6}{$6$ cusps}  
  \includegraphics[scale=0.35]{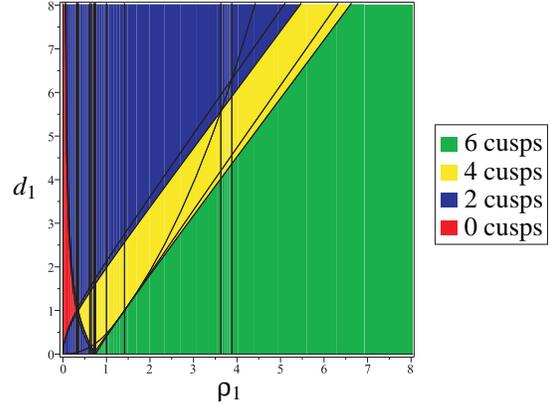}
  \caption{Cell Decomposition for $(\rho_1,d_1)$}
  \label{Fig_cellsr1d1_4colors}
\end{figure}
\begin{figure}[t]
  \psfrag{r1}{$\rho_1$}
  \psfrag{d1}{$d_1$}
  \includegraphics[scale=0.4]{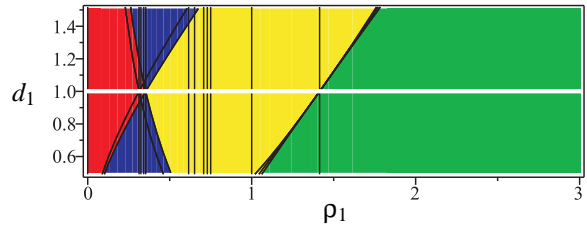} 
  \caption{\label{Fig_cellsr1d1_zoom}Zoom in of the cell decomposition for $(\rho_1,d_1)$. Line $d_1=1$ in white. }
\end{figure}

The complement of this DV produces 90 cells with associated numbers of cusps
varying among 0, 2, 4 and 6, as shown in Fig.~\ref{Fig_cellsr1d1_4colors}, where
 black vertical lines  delimit intersection points of the DV. Let us notice that
although all cells must be considered disconnected, it is apparent that cells
are naturally grouped with regard to their number of cusps. Additionally, cells 
 with different number of cusps are exclusively separated by curves of the
discriminant variety.  Furthermore, this distribution is consistent with that
obtained for $d_1=1$ in Section~\ref{subsec_CaseStudyDiscussion}, as can be seen
in Fig.~\ref{Fig_cellsr1d1_zoom}. However, here the section is divided into many
more smaller intervals whose borders we cannot apriori ensure to be associated
to a specific number of cusp points. In fact, let us also observe that the cells
with 2 cusp points (in blue) degenerate into one single point for $d_1=1$. So we
could claim that the case with $d_1=1$  is a very special degenerate
3-R\underline{P}R manipulator.   

%%%%%%%%%%%%%%%%%%%%%%%%%%%%%%%%%%%%%%%%%%%%%%%%%%%%%%%%%%%%%%%%%%%%%%%%%
\subsection{Study of the cusp points on the borders of the CAD \label{subsec_borders}}
%%%%%%%%%%%%%%%%%%%%%%%%%%%%%%%%%%%%%%%%%%%%%%%%%%%%%%%%%%%%%%%%%%%%%%%%%
At this point we can only certify the number of cusp points in the open cells.
This excludes the cell borders. The union of all these borders consists of the
DV plus the delimitation of intersection points of the DV. However, by
definition, changes in the number of solutions can only happen on the DV. So, we
just need to analyze the DV.

We could try executing a further iteration of the CAD on the DV, but the system
to be solved turns out to be too complex and we cannot obtain any results after
long computations.  It is clear that not all points on the DV will correspond to
the same number of solutions.  But we can expect the number of solutions to be
preserved along the DV   between two consecutive auto-intersection points of the
DV.  And although it has not yet been proven, many tests have been run on
several examples with random points on the DV and all the results confirm what
the following conjecture infers. 

\noindent\textbf{Conjecture 1.} Given a polynomial system $F$, let $\mathcal{V}$
be the DV of $F$ w.r.t two parameters $U_1, U_2$, and let $\mathcal{A}$ be the
set of its auto-intersection points. Then, the number of solutions of $F$ is
constant on each connected component of $\mathcal{V} \setminus \mathcal{A}$.

The following algorithm analyzes the to study the numbers of solutions on the DV
based on the previous conjecture.
\begin{algorithm}[H]
\caption{Number of solutions of $F$ on the DV}
\label{Alg_cellborders}
\begin{algorithmic}
\STATE{$\mathcal{V}=$ variety of $DV(F;U_1,U_2)$} 
\STATE{$\mathcal{A}=$ $\{$auto-intsersection points of $\mathcal{V}\}$} 
\FOR{each connected component $\mathcal{U}_i$ of $\mathcal{V} \setminus \mathcal{A}$}
  \STATE{$p_i=$ random point on $\mathcal{U}_i$} 
  \STATE{Compute the number of solutions on $\mathcal{U}_i$ as \\ the number of solutions of $F|_{p_i}$}
\ENDFOR
\FOR{each point $q \in \mathcal{A}$}
  \STATE{Compute the number of solutions of $F|_{q}$}
\ENDFOR
\end{algorithmic}
\end{algorithm}

%%%%%%%%%%%%%%%%%%%%%%%%%%%%%%%%%%%%%%%%%%%%%%%%%%%%%%%%%%%%%%%%%%%%%%%%%
\subsection{Complete analysis and applications \label{subsec_CompleteDiscussionPractice}}
%%%%%%%%%%%%%%%%%%%%%%%%%%%%%%%%%%%%%%%%%%%%%%%%%%%%%%%%%%%%%%%%%%%%%%%%%
From the results exposed in the previous subsections and by joining all the
different pieces together we obtain a complete partition of the 2-dimensional
parameter space. 

\begin{figure}[t]
\begin{minipage}[b]{\linewidth}
\centering
\psset{unit=0.75cm}
\begin{pspicture}(0.1,0.9)(11,2.6)
  \psframe[fillstyle=solid,fillcolor=red, linecolor=red] (0.2,2) (0.8,2.5)
  \psframe[fillstyle=solid,fillcolor=yellow, linecolor=yellow] (0.2,1) (0.8,1.5)
    \rput[l]{0}(1.05,2.2){\em $0$ cusps}
    \rput[l]{0}(1.05,1.2){\em $4$ cusps}

  \psframe[fillstyle=solid,fillcolor=cyan, linecolor=cyan] (3,2) (3.6,2.5)
  \psframe[fillstyle=solid,fillcolor=gray, linecolor=gray] (3,1) (3.6,1.5)
    \rput[l]{0}(3.85,2.2){\em $1$ cusps}
    \rput[l]{0}(3.85,1.2){\em $5$ cusps}
    
  \psframe[fillstyle=solid,fillcolor=blue, linecolor=blue] (5.8,2) (6.4,2.5)
  \psframe[fillstyle=solid,fillcolor=green, linecolor=green] (5.8,1) (6.4,1.5)
    \rput[l]{0}(6.65,2.2){\em $2$ cusps}
    \rput[l]{0}(6.65,1.2){\em $6$ cusps}
    
  \psframe[fillstyle=solid,fillcolor=black, linecolor=black] (8.6,2) (9.2,2.5)    
    \rput[l]{0}(9.45,2.2){\em $3$ cusps}
 \psframe[linecolor=black, linewidth=0.5pt] (0.1,0.9) (11,2.6)    
\end{pspicture}
\end{minipage}
\vspace{0.1mm}

\begin{minipage}[b]{0.45\linewidth}
\centering
   \psfrag{r1}{$\rho_1$}
   \psfrag{d1}{$d_1$}
   \includegraphics[scale=0.21]{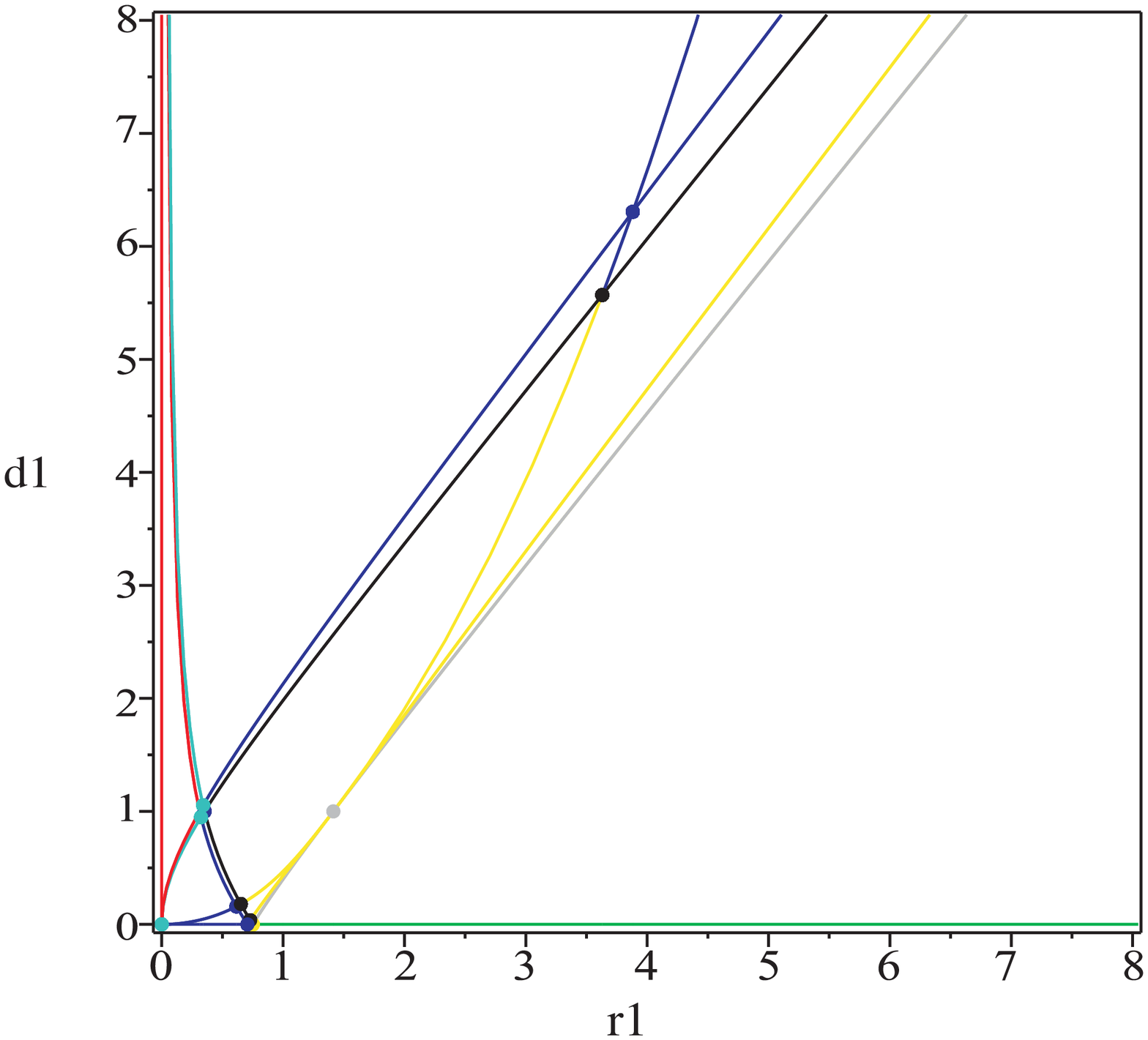}
\end{minipage}
\hspace{0.04\linewidth}
\begin{minipage}[b]{0.45\linewidth}
\centering
   \psfrag{r1}{$\rho_1$}
   \psfrag{d1}{$d_1$}
   \includegraphics[scale=0.203]{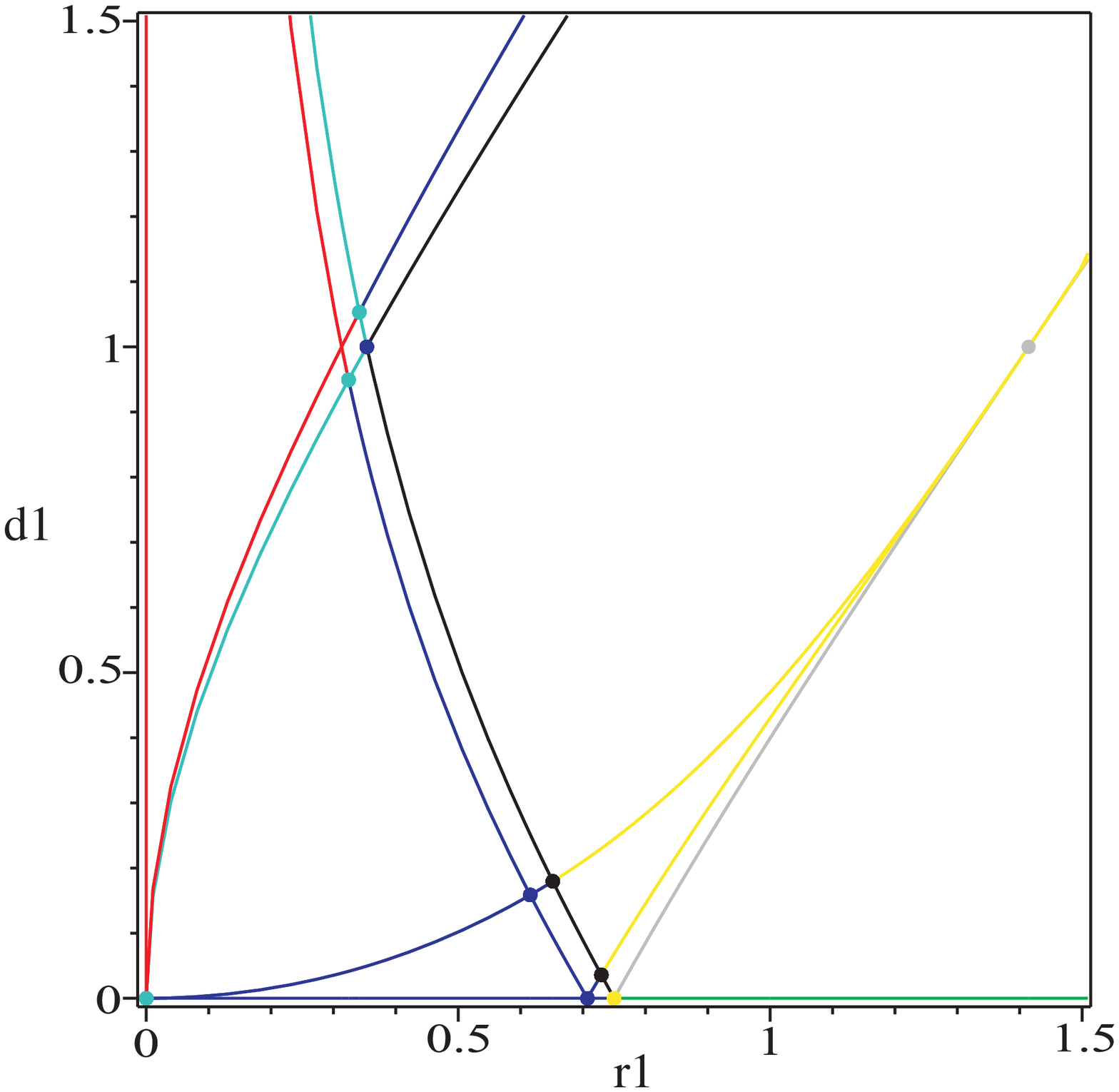}
\end{minipage}
\begin{center}
(a) \hspace{3.7cm} (b) 
\end{center}
\caption{\label{Fig_DVcolorspoints_r1d1}
Distribution of cusp points on $DV(C_F;\rho_1,d_1)$ (a), and zoom in view on $[0,1.5]\times[0,1.5]$ (b).
}
\end{figure}

\begin{figure}[t]
\centering
   \psfrag{r1}{$\rho_1$}
   \psfrag{d1}{$d_1$}
   \includegraphics[scale=0.3]{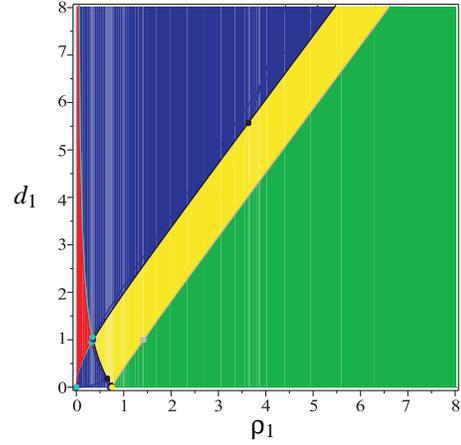}
   \caption{Complete analysis of the cusp points for $(\rho_1,d_1)$}
   \label{Fig_cellsd1r1_complete}
\end{figure}

The execution of Algorithm~\ref{Alg_cellborders} on $DV(C_F; \rho_1, d_1)$
provides the distribution of cusp points shown in
Fig.~\ref{Fig_DVcolorspoints_r1d1}. The integral picture of the 2-dimensional
distribution is given in Fig.~\ref{Fig_cellsd1r1_complete}. It is interesting to
notice that there is a continuity on the transitions between cells having the
same number of cusp points, since their common border inherits that same number
of cusp points.

Observe also that this distribution   has been obtained thanks to the DV
associated to the chosen parameters $d_1$ and $\rho_1$, which  depends
exclusively on these two parameters. In particular, $DV(C_F;\rho_1,d_1)$ does
not depend on $\rho_2$ nor $\rho_3$. This tends to be erroneously  interpreted
as: \\
 ``\emph{if we pick a $(\rho_1,d_1)$-point with associated number of cusps $k$,
then whatever the values $\rho_2$ and $\rho_3$ may take,
$C_F|_{(\rho_1,\rho_2,\rho_3, d_1)}$ has $k$ solutions}''.\\
Instead, it should be read as follows: \\
``\emph{if we pick a $(\rho_1,d_1)$-point with associated number of cusps $k$
and fix these values  then, among the reachable configurations there are $k$
cuspidal ones, i.e. $C_F|_{(\rho_1, d_1)}$ has $k$ solutions}''. 
However, the number of associated cusp points does establish a maximum of
cuspidal configurations for any values $\rho_2$ and $\rho_3$. For example, in
yellow regions we can have a maximum of $4$ cuspidal configurations, but
depending on the values of $\rho_2$ or $\rho_3$ there can even be none. In
particular, for the red regions there are $0$ cusp points for all possible
values $\rho_2$ and $\rho_3$.

\begin{figure}[t]
\centering
   \psfrag{r1}{$\rho_1$}
   \psfrag{d1}{$d_1$}
   \psfrag{r1n}{$\Delta\rho_{1}$}
   \psfrag{r1m}{ }
   \includegraphics[scale=0.4]{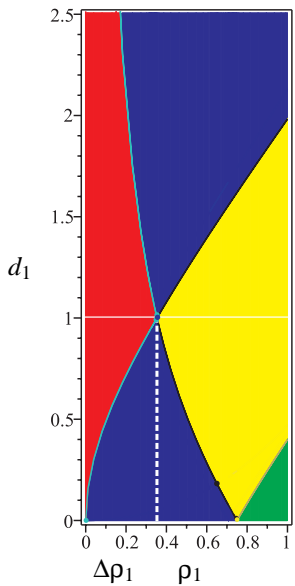}
   \caption{Optimal $d_1$ and $\Delta \rho_1$ for non-cuspidal degenerate 3-R\underline{P}R.}
\label{Fig_cellsd1r1_complete_zoom_fixr1_noncuspidal}
\end{figure}

Some applications can be derived which may be interesting  
from the designer's point of view: 
\begin{itemize}
\item[$\bullet$] It can be helpful in deciding the most suitable architecture of
the mechanism. Let us assume that we want to design a 3-R\underline{P}R
manipulator with some given geometric constraints such that for a specific task
one of the legs has to be blocked to a fixed length $\rho_1$,  but the job
requires a large singularity-free workspace. Therefore, we may be interested in
finding a range $\Delta d_1$ of parameter values for which the manipulator is
cuspidal.
\item[$\bullet$] It can also be useful for deciding the most suitable ranges of
leg lenghts for each possible architecture, given a specific task. For instance,
let us assume that the job is set for a non-cuspidal manipulator with parameter
values $A_{2x}=1$, $A_{3x}=0$, $A_{3y}=1$, $\beta=-\pi/2$, and $d_3=1$, but it
requires the largest possible range of the leg length $\rho_1$. Then, the value
$d_1$ can be optimized with this criterion.
Figure~\ref{Fig_cellsd1r1_complete_zoom_fixr1_noncuspidal} details both the
optimal valule $d_1$ and the largest possible range $\Delta \rho_1$ for our
problem. 
\end{itemize}

Let us just notice that in both cases the obtained ranges  can be of varied
topology (open, closed, semi-closed, open and closed, connected, or even a union
of these types). This is due to the combination of both the CAD and the study of
the cusp locus on the DV.

%%%%%%%%%%%%%%%%%%%%%%%%%%%%%%%%%%%%%%%%%%%%%%%%%%%%%%%%%%%%%%%%%%%%%%%%%
%%%%%%%%%%%%%%%%%%%%%%%%%%%%%%%%%%%%%%%%%%%%%%%%%%%%%%%%%%%%%%%%%%%%%%%%%
\section{Conclusions \label{sec_conclusion}}
%%%%%%%%%%%%%%%%%%%%%%%%%%%%%%%%%%%%%%%%%%%%%%%%%%%%%%%%%%%%%%%%%%%%%%%%%
%%%%%%%%%%%%%%%%%%%%%%%%%%%%%%%%%%%%%%%%%%%%%%%%%%%%%%%%%%%%%%%%%%%%%%%%%
This paper has introduced both an efficient method for the computation of the
cuspidal configurations of a mechanism, and a reliable algorithm that partitions
a given parameter space into open regions with constant number of associated
cusp points. 

The first one is based on a symbolic-algebraic approach able to describe the
roots of exact multiplicity $3$ and a certified numerical algorithm that
isolates among them the real (i.e. not complex) ones. This symbolic-numeric
approach is more efficient than other previously existing methods, which mainly
relied on the approximation of roots of multiplicity at least $3$ after reducing
the initial system to a simpler one and projecting it onto the $\bm\rho$-space.

This new method is combined with some algebraic tools such as the discriminant
variety (DV) and the cylindric algebraic decomposition (CAD) in order to analyze
a 2-dimensional parameter space with respect to the associated number of cusp
points. This second algorithm provides a partition of the parameter space into
cells with constant number of cusp points, which is  certified for whatever
values are picked inside each open cell but not on their borders. Cell borders
are further analyzed by Algorithm~\ref{Alg_cellborders} based on Conjecture 1,
which still remains unproved.

Both algorithms have been applied to the analysis and distribution of the cusp
locus for a family of degenerate 3-R\underline{P}R manipulators,  and some
applications to robot design are also derived.  This does not mean that the two
given algorithms are specially designed for this type of 3-R\underline{P}R
mechanisms. Indeed, they are suitable for more general examples
since they do not rely on any ad-hoc formulation. Nevertheless for some examples
the obtention of results within reasonable time may not be feasible yet, since
there is an important symbolic-algebraic part,  and thus the more complex the
initial system is the harder it will be to compute a partition on the parameter
space. 

%%%%%%%%%%%%%%%%%%%%%%%%%%%%%%%%%%%%%%%%%%%%%%%%%%%%%%%%%%%%%%%%%%%%%%%%%
%%%%%%%%%%%%%%%%%%%%%%%%%%%%%%%%%%%%%%%%%%%%%%%%%%%%%%%%%%%%%%%%%%%%%%%%%
\section*{Acknowledgements}
%%%%%%%%%%%%%%%%%%%%%%%%%%%%%%%%%%%%%%%%%%%%%%%%%%%%%%%%%%%%%%%%%%%%%%%%%
%%%%%%%%%%%%%%%%%%%%%%%%%%%%%%%%%%%%%%%%%%%%%%%%%%%%%%%%%%%%%%%%%%%%%%%%%
The authors would like to acknowledge the financial support of this research by
ANR Project SiRoPa. The first and second authors have been supported for this
research by two postdoctoral contracts at the Institut de Recherche en
Communications et Cybern\'etique de Nantes under the sponsorship of the same
project.

%%%%%%%%%%%%%%%%%%%%%%%%%%%%%%%%%%%%%%%%%%%%%%%%%%%%%%%%%%%%%%%%%%%%%%%%%
%%%%%%%%%%%%%%%%%%%%%%%%%%%%%%%%%%%%%%%%%%%%%%%%%%%%%%%%%%%%%%%%%%%%%%%%%
\bibliographystyle{asmems4}
\bibliography{ManubensMorozChablatWengerRouillier}
%%%%%%%%%%%%%%%%%%%%%%%%%%%%%%%%%%%%%%%%%%%%%%%%%%%%%%%%%%%%%%%%%%%%%%%%%
%%%%%%%%%%%%%%%%%%%%%%%%%%%%%%%%%%%%%%%%%%%%%%%%%%%%%%%%%%%%%%%%%%%%%%%%%
\end{document}